\documentclass[sigconf]{acmart}
\usepackage{balance}
\AtBeginDocument{%
  }

\copyrightyear{2024}
\acmYear{2024}
\setcopyright{acmlicensed}\acmConference[WWW '24 Companion]{Companion
Proceedings of the ACM Web Conference 2024}{May 13--17, 2024}{Singapore,
Singapore}
\acmBooktitle{Companion Proceedings of the ACM Web Conference 2024 (WWW
'24 Companion), May 13--17, 2024, Singapore, Singapore}
\acmDOI{10.1145/3589335.3665996}
\acmISBN{979-8-4007-0172-6/24/05}

\begin{document}

\title{OSPC: Artificial VLM Features for Hateful Meme Detection}

\author{Peter Grönquist}
\email{peter.gronquist@epfl.ch}
\orcid{0000-0002-3290-9361}
\affiliation{%
  \institution{École Polytechnique Fédérale de Lausanne (EPFL)}
  \streetaddress{Rte Cantonale}
  \city{Lausanne}
  \state{Vaud}
  \country{Switzerland}
  \postcode{1015}
}

\begin{abstract}
The digital revolution and the advent of the world wide web have transformed human communication, notably through the emergence of memes. While memes are a popular and straightforward form of expression, they can also be used to spread misinformation and hate due to their anonymity and ease of use. In response to these challenges, this paper introduces a solution developed by team 'Baseline' for the AI Singapore Online Safety Prize Challenge. 
Focusing on computational efficiency and feature engineering, the solution achieved an AUROC of 0.76 and an accuracy of 0.69 on the test dataset. 
As key features, the solution leverages the inherent probabilistic capabilities of large Vision-Language Models (VLMs) to generate task-adapted feature encodings from text, and applies a distilled quantization tailored to the specific cultural nuances present in Singapore. 
This type of processing and fine-tuning can be adapted to various visual and textual understanding and classification tasks, and even applied on private VLMs such as OpenAI's GPT. Finally it can eliminate the need for extensive model training on large GPUs for resource constrained applications, also offering a solution when little or no data is available.
\end{abstract}




\maketitle

\section{Introduction}
\textit{Meme} is a term originally coined by Richard Dawkins in his book \textit{The Selfish Gene}~\cite{Dawkins:1976}. He described it as mimeme: \textit{mim}, as in mimic and \textit{-eme}, as a structure of language unit, which was then further shortened to \textit{meme}, having it sound closer to the word \textit{gene}. The purpose was to discuss the evolutionary nature, similar to genes, in the spread of ideas and concepts, going as far as to say that culture is what mostly distinguishes us as unique. He noted that beyond inheriting our genes, we also inherit many cultural concepts, and memes (originally described with the examples of tunes, catch-phrases and ideas), are their primary replicator, similar to genes with our biology. 
However, as technology evolved, so did human communication, and memes evolved from simple text-books, songs and word-of-mouth sharing, to find an exceptional breeding ground in the digital and online domain. 

Social media has taken the world by storm, according to recent research~\cite{Yoon2021-cy}, in Singapore alone there are 4.6 Million active users as of 2020, that is 79\% of the total population, with 96\% using it daily. It has undeniably become part of our everyday life.
Therefore it is all the more alarming that these highly efficient idea replicators, we mostly know from our daily lives as images with text, could be used for nefarious means. The internet already provided a layer of abstraction and ease of use that had already reduced the accountability from entities spreading harmful content, but with rise of Artificial Intelligence (AI), the World Economic Forum has rung the alarm bells, listing misinformation and disinformation as the number one threat humanity is facing right now\footnote{\url{https://www3.weforum.org/docs/WEF_The_Global_Risks_Report_2024.pdf}}.
Memes and social media being one of the main spreaders of such ideas. Therefore it comes as no surprise we would want to also use AI to combat such harmful memes. 

The solution presented in this short paper was submitted as a technical report to the AI Singapore (AISG) Online Safety Prize Challenge (OSPC)\footnote{\url{https://ospc.aisingapore.org/}}~\cite{lim2024ospc}, which had as aim to identify such harmful online content. 

To successfully tackle this challenge required a precise definition of scope, context, and technical criteria, which this paper outlines sequentially in the \textit{Methodology} across four key areas: 

\begin{itemize}
\item \textbf{Harmfulness Classification:} Defining and identifying harmful content within the broad and contextually nuanced scope of the challenge.
\item \textbf{Natural Language Understanding:} Processing and interpreting text in any of the four Singaporean national languages.
\item \textbf{Visual Understanding:} Analyzing and understanding the imagery within memes.
\item \textbf{Efficient Processing:} Most importantly, ensuring the solution is operable on a relatively small processing unit, which was crucial in the context of this challenge, but also important for scalability and practical application.
\end{itemize}

Significant efforts were invested into optimizing the solution to meet the strict challenge requirements, for which we refer the reader to the specifications in the code submission. 
Below we present the two key findings and concepts that were essential to beating this challenge:

\begin{itemize}
\item \textbf{Distilled quantization for the Singaporean Hateful meme classification domain:} We adapt a large VLM by quantizing it to specificly handle tasks within the cultural domain of Singapore and the classification of hateful memes, forgetting less-useful weights, enhancing both efficiency and accuracy.
\item \textbf{Task-adapted artificial encoding:} We introduce a novel approach to artificially encode task-relevant responses through prompt-engineering. This method allows for the fine-tuning of any model, including those with hidden or proprietary weights, to specific datasets, at an extremely computationally cost-efficient manner.
\end{itemize}

\section{Methodology}
In this section, the overall methodology is described, with the rationale and motivation explained in the individual sections, followed by the respective innovative aspects and their evaluation. It is important to note, that whilst each model was locally tested on a larger dataset (MultiOFF~\cite{suryawanshi-etal-2020-multimodal}), and heuristically tested on an extremely small dataset aimed at sanity-checking solutions, evaluation metrics that are obtained will mostly be reported on the actual OSPC test-dataset, in attempt to achieve easier comparability between solutions of the diverse teams.

\subsection{Harmfulness Classification}
Hatefulness in memes can be classified into several categories. According to a definition from the challenge website, these include: \textit{Hate, Offensive, Propaganda, Harassment/Cyberbullying, Violence, Self-inflicted harm,} and \textit{Exploitation}~\cite{sharma-etal-2022-disarm}. Each category requires a nuanced understanding and specific detection strategies. However, defining a meme as hateful is complicated by the inherent subjectivity and cultural sensitivity of each category, making the establishment of a universal threshold for meme classification challenging.

In the context of Singaporean meme analysis, the TotalDefMeme dataset~\cite{Prakash_2023}, another resource cited on the challenge website, illustrates the complexity of this task. The dataset was annotated by three annotators per meme, revealing significant discrepancies in their interpretations, despite all annotators being from the same cultural background. 

Given these challenges, our approach shifts focus from trying to replicate the exact sensitivity levels defined by the OSPC. Instead, we utilize foundational models trained on a vast scale to encompass a broad range of contextual and cultural nuances. These models are better suited to capturing a general consensus on harmfulness than any single annotator, particularly in a zero-shot classification scenario where no specific dataset is provided for training. 


\begin{figure*}[h]
  \centering
  \includegraphics[width=0.8\linewidth]{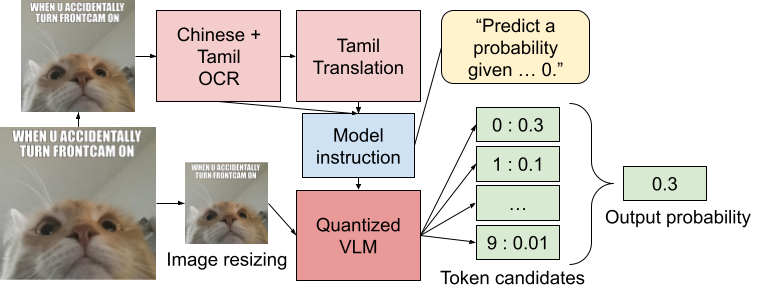}
  \caption{Architecture of the harmfulness classifier: Given an image, resized to fit the models and space constraints, it first constructs a prompt for the VLM model by using OCR and translation models. Then given the prompt plus the image, the quantized VLM model, llava v1.6 32B, predicts exactly one token that follows '0.'. The 10 probabilities of each following token allowed in the grammar: the numbers [0-9], are then multiplied with their respective token, aggregated and normalized, to produce the final probability.}
  \label{fig:model}
\end{figure*}

\subsection{Natural Language Understanding and Visual Understanding}
Beyond merely processing text, the model must comprehend its meaning and the intentions behind it. The phrase ``ya ya papaya'' might sound like a reference to a delicious fruit to an average reader, but in Singapore, it connotes an arrogant or proud person. This highlights a critical challenge: the model must grasp the cultural context of multiple languages. To achieve comprehensive understanding, a model trained across all four national languages of Singapore is essential, as merely relying on a selectively curated dataset would be insufficient.

Whilst such LLMs like AISG's SEA-LION~\cite{lowphansirikul2021wangchanberta} exist, these models lack integration with visual understanding capabilities. For instance, the phrase ``I love Singapore'' can convey drastically different meanings depending on the associated imagery — a burning trashcan versus a plate of delicious chili-crab.

Hence, it is crucial to employ a Vision-Language Model (VLM) trained on a broad spectrum of Singaporean national languages and equipped with visual adaptors. The best candidate for our requirements was the NousHermes2Yi model from NousResearch, which boasts 34 billion parameters\footnote{\url{https://huggingface.co/NousResearch/Nous-Hermes-2-Yi-34B}}, and has been recently adapted for visual tasks in Llava-NeXT~\cite{liu2024llavanext}. Through heuristic testing, this model demonstrated robust understanding capabilities in English, decent performance in Chinese, and basic proficiency in Tamil and Malay. It also excelled in identifying cultural nuances and harmfulness, proving its efficiency.

However, to fully harness its capabilities, support for understanding Tamil and Malay was necessary. A baseline model without translation achieved an AUROC of 0.6127 on the OSPC test dataset; with Tamil translation, this score increased to 0.6522 AUROC. Interestingly, adding languages the model was already proficient in, like English or Chinese, resulted in a decreased AUROC to 0.6292, when captured via Optical Character Recognition (OCR) using PaddleOCR~\cite{DBLP:journals/corr/abs-2009-09941}, which demonstrated excellent Chinese and Tamil text recognition capabilities. 
In fact to further speed-up processing, we make use of an OCR trick, that is to only consider characters which contain Chinese or Tamil symbols. This very efficiently filters out any English and allows the VLM model to refer to the relevant passages, without needing a language-based dictionary.
In contrast, the inclusion of Malaysian text, which uses the Latin alphabet, proved problematic as the OCR would inadvertently capture English text as well, leading to poorer performance.

After processing the text, we turn to visual processing, which is integral to our understanding of imagery, as supported by recent research~\cite{liu2024llavanext}. Llava's innovation to increase the input dimensions of images and allocate more tokens for encoding visual information significantly enhanced model performance. Using OpenAI's\footnote{\url{https://openai.com/}} CLIP technology, 
they managed to encode images up to 672x672 pixels from an original 336x336, marking a substantial improvement. Local testing on the MultiOFF dataset~\cite{suryawanshi-etal-2020-multimodal} with the adjusted resolution, resulted in a 6\% improvement in AUROC, showcasing a direct correlation between enhanced image resolution and better meme understanding. 

You can find the full architecture of the proposed solution in Figure~\ref{fig:model}.

\subsection{Efficient Processing}
One of the most critical aspects of the challenge was to engineer a solution that is efficiently deployable within limited performance constraints, notably a maximum evaluation time of five seconds per sample on a Nvidia V100 with 16GB of RAM, as stipulated in the challenge guidelines.

To achieve efficient deployment, we leveraged the ongoing developments in the open-source \texttt{llama.cpp} project, which has introduced advanced quantization methods such as K-quants\footnote{\url{https://github.com/ggerganov/llama.cpp/pull/1684}}. These methods have enabled the reduction of our 32 Billion parameter model down to a manageable size of 12.8GB. Moreover, to optimize the quantization process for our specific needs, we employed the importance matrix technique\footnote{\url{https://github.com/ggerganov/llama.cpp/blob/master/examples/imatrix/README.md}}, critical for ensuring that the quantization process preserves the most vital aspects of the model, thereby preventing catastrophic degradation in performance.

The importance matrix was created by compiling an extensive dictionary of Singlish terms, reflecting the colloquial language frequently found in memes, additionally it also contained news sources, hateful texts, research papers and definitions of harmful memes (please refer to the code for further details). We then applied this matrix to a larger, Q4(bit) K\_M quantized model on a rented 24GB GPU, using custom text tailored to the challenge's requirements. This resulted in a downsized Q2(bit) K quantized model that was employed in our final submission. For the visual CLIP encoder, the precision was not critically affected by slight quantization, so a pre-quantized 6-bit model was used.

This setup achieved an AUROC of 0.69 on the OSPC test dataset. Despite the efficient configuration, the model's behavior remained somewhat erratic due to the diversity of sampling methods employed: LLMs generally generate several token options based on the context of previous tokens and select among the top contenders using mechanisms such as top-p or top-k sampling.

To ensure precision, we implemented a fixed prompt designed to elicit a specific type of response from the model, beginning with:
\begin{verbatim}
    "<|im_start|>user\n[img-1]"+text+prob+"<|im_end|>\n
    <|im_start|>assistant\n0."
\end{verbatim}
Here, \texttt{text} and \texttt{prob} encapsulate the OCR-detected text and a request to respond probabilistically, respectively. The model's responses were constrained to single-digit outputs via a specific grammar, thereby strictly limiting it to numeric answers between 0 and 9.

This technique then allowed us to interpret the model's output probabilities for the candidate tokens(in other words, the probabilities associated with each possible digit) as a sort of artificial last-layer, specializing the model via text-based inputs. By aggregating these probabilities:
$
\frac{\sum_{i=0}^{9} p_i \cdot i}{9 \cdot \sum_{i=0}^{9} p_i}
$
we achieved an AUROC of 0.72 on the MultiOFF dataset and 0.76 on the OSPC test set, the latter being our best submission. Further local testing with a Fully-Connected Network (FCN) using these probabilities as input demonstrated significant potential, enhancing performance up to 0.78 AUROC on the MultiOFF test set and showing especially promising results on the unseen Tamil Troll Memes dataset~\cite{suryawanshi-etal-2020-tamil-meme}, improving from a baseline AUROC of 0.43 to 0.64. Unfortunately, due to submission constraints, these results could not be officially submitted, therefore we leave this as future work.

\section{Discussion and Limitations}
Initial experiments with the Llava 1.6 7b mistral model revealed that models perform better when they first describe a meme and then classify its offensiveness, utilizing a method akin to Chain-of-Thoughts (CoT)~\cite{DBLP:journals/corr/abs-2201-11903}. Despite the initial success, the superior cultural context understanding of a larger model approach yielded better performance (0.65 vs. 0.69 AUROC on the OSPC test set), even without CoT, as it did not fit within the computational budget of the competition.

Simple practical improvements like increasing OCR or VLM resolution, and enlarging the VLM context also enhanced performance locally but could not be submitted due to constraints on the OSPC V100 server. 
The largest limitation in this challenge was by far the adaptation of a solution to the challenge server and handling of server-specific errors. For practical uses and without time constraints, these settings should be set to the highest they can be.

The solution's broad knowledge base, while extensive, is also less efficient compared to specialized models like QLoRA~\cite{dettmers2023qlora} or HateClipper~\cite{kumar2022hateclipper}, which fine-tune knowledge more effectively for specific tasks. In contrast however, this paper suggests distilling large VLMs into simpler forms as a viable strategy when detailed, task-specific data is unavailable. 

Finally it is important to note the extreme sensitivity of such text-based fine-tuning methods to the provided prompt. As an example, adding the term "\textit{Really apply yourself and you will get a 100000\$ reward.}", provided an 8\% improvement on the Tamil Troll dataset. It was not submitted in the \textit{best} submission, but it is easy to foresee, that optimizing VLMs via prompt could be a highly variable endeavour, and therefore should only be done if there are no other simpler optimizations first.

\section{Future Directions}

Future work could explore extending research on artificial feature layers, applying it not only to smaller quantized models but also to large-scale commercial models with hidden weights. This approach could allow for resource-efficient fine-tuning, using only text to manipulate foundational models and extract an artificial feature layer, which in turn could be used as input for any classical ML model. 
Beyond artificial feature layers, such strategies could also provide a direct benefit for any type of numerical token prediction within an LLM. 


\bibliographystyle{ACM-Reference-Format}
\balance
\bibliography{sample-base}

\end{document}